\documentclass[letterpaper, 10 pt, conference]{ieeeconf}  

\let\labelindent\relax
\usepackage[inline]{enumitem}
\usepackage{subcaption}
\usepackage{epsfig}
\usepackage{dblfloatfix}
\usepackage{float}
\usepackage{hyperref} 
\usepackage{amssymb}
\usepackage{amsmath,amsfonts}
\usepackage{multicol}
\usepackage{multirow}

\IEEEoverridecommandlockouts                              

\title{\LARGE \bf
End-to-End Intersection Handling using Multi-Agent Deep Reinforcement Learning
}

\author{Alessandro Paolo Capasso$^{1}$, Paolo Maramotti$^{1}$, Anthony Dell'Eva$^{2}$ and Alberto Broggi$^{3}$
\thanks{$^{1}$Alessandro Paolo Capasso, $^{1}$Paolo Maramotti are with VisLab - University of Parma, Parma, Italy
        {\tt\small alessandro.capasso@unipr.it, paolo.maramotti@unipr.it}}%
\thanks{$^{2}$Anthony Dell'Eva is with VisLab - University of Bologna, Bologna, Italy 
		{\tt\small anthony.delleva2@unibo.it}}
\thanks{$^{3}$Alberto Broggi is with VisLab srl, an Ambarella Inc. company - Parma, Italy
        {\tt\small broggi@vislab.it}}%
}

\begin{document}

\maketitle
\thispagestyle{empty}
\pagestyle{empty}

\begin{abstract}

Navigating through intersections is one of the main challenging tasks for an autonomous vehicle. However, for the majority of intersections regulated by traffic lights, the problem could be solved by a simple rule-based method in which the autonomous vehicle behavior is closely related to the traffic light states. In this work, we focus on the implementation of a system able to navigate through intersections where only traffic signs are provided. We propose a multi-agent system using a continuous, model-free Deep Reinforcement Learning algorithm used to train a neural network for predicting both the acceleration and the steering angle at each time step. We demonstrate that agents learn both the basic rules needed to handle intersections by understanding the priorities of other learners inside the environment, and to drive safely along their paths. Moreover, a comparison between our system and a rule-based method proves that our model achieves better results especially with dense traffic conditions. Finally, we test our system on real world scenarios using real recorded traffic data, proving that our module is able to generalize both to unseen environments and to different traffic conditions.

\end{abstract}

\section{INTRODUCTION}

The use of Deep Reinforcement Learning (DRL)~\cite{DRL} algorithms is growing exponentially, from the resolution of games like Atari~\cite{atari} and Go~\cite{go} to the robotic field~(\cite{robotic1},~\cite{robotic2},~\cite{mujoco}). Indeed, these algorithms have proved to achieve impressive results both in discrete control-space problems~\cite{discrete} and in continuous ones~(\cite{continuous1},~\cite{continuous2}). In the last decade, DRL algorithms have also been used in the autonomous driving field to solve many control tasks like lane change~\cite{lanechange}, lane keeping~\cite{lanekeeping}, overtaking maneuvers~\cite{overtaking} and many others.

However, many situations are not so trivial to handle by an autonomous vehicle; for example, navigating through an intersection may be a difficult task in absence of traffic lights or with heavy traffic conditions. Indeed, in most intersections regulated by traffic lights, this problem could be solved by a simple rule-based approach in which the autonomous car behavior strictly depends on the traffic light state. The typical solution used for dealing with those cases in which traffic lights are not present are based on time-to-collision algorithm (TTC)~\cite{ttc}, that could be useful for simple cases but it has several limitations: it assumes costant speed of traffic vehicles, it does not understand the dynamic of the scenario and the possible intentions of other agents and it could lead to unnecessary delays.

\begin{figure*}[h]
\vspace{0.3cm}
  \centering
  \begin{subfigure}{.24\linewidth}
    \centering
    \includegraphics[width =\linewidth]{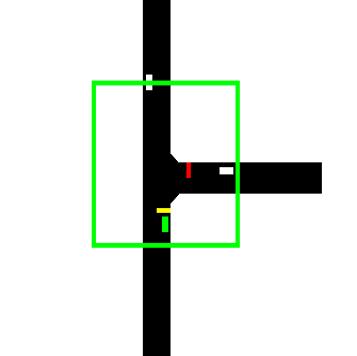}    
    \caption{Easy intersection}
    \label{fig:pharmacy}
  \end{subfigure}
  \begin{subfigure}{.24\linewidth}
    \centering
    \includegraphics[width =\linewidth]{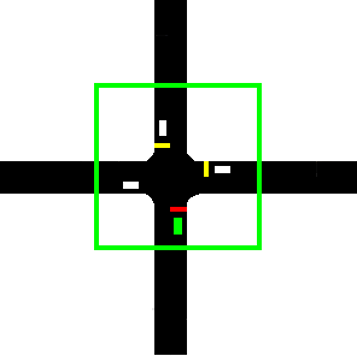}
    \caption{Medium intersection}
    \label{fig:cross4_sl}
  \end{subfigure}
  \begin{subfigure}{.24\linewidth}
    \centering
    \includegraphics[width =\linewidth]{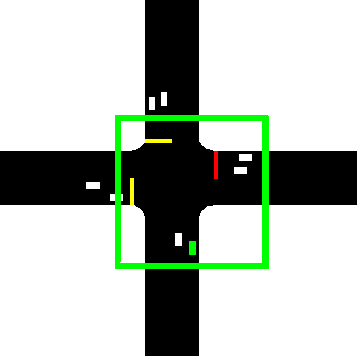}
    \caption{Hard intersection}
    \label{fig:cross4}
  \end{subfigure}
  \caption{Synthetic representations of three intersections representing three different difficulty levels, from the easiest (Fig.~\ref{fig:pharmacy}) to the hardest one (Fig.~\ref{fig:cross4}). Each agent in the scenario observes an area of 50$\times$50 meters: green squares show some examples of these surroundings perceived by green vehicles. }
  \label{fig:synthetic_intersections}
\end{figure*}

In this paper, we focus on intersections regulated by traffic signs in which agents are trained in a multi-agent fashion. Vehicles learn to handle such scenarios learning the priority to the right rule, that is a right of way system widely used in countries with right-hand traffic. In this way, when two vehicles are approaching the intersection and their trajectories may intersect, the car coming from the right has the priority in the case both agents have the same traffic sign or when no sign is present; otherwise, the priority is defined by traffic signs. Together with this task, agents learn to drive along their paths through the training of a neural network that predicts the acceleration and the steering angle at each time step. The whole training is done in simulation as the majority of the DRL algorithms; we use a synthetic representation of real intersections~(Fig.~\ref{fig:synthetic_intersections}), developing our simulator with CAIRO graphic library~\cite{cairo} in order to reduce the state complexity respect to realistic graphic simulators like CARLA~\cite{carla} or GTA-based platforms (\cite{gta1},~\cite{gta2}). In this way, the transfer of a policy between synthetic and real domain could be easier to achieve as explained in~\cite{simtoreal}, even if recent works have shown encouraging results for domain transfer between simulated and real-world images~\cite{kendall2}.

In this paper we propose a system in which agents learn the right of way rule in a multi-agent fashion to handle intersection scenarios. Moreover, we demonstrate that our approach reaches better performance than a rule-based method, especially with dense traffic conditions, where a conservative approach leads the system to undefined waits. We also show that our model features generalization capabilities both for the safe execution of the crossing maneuver and for a safe end-to-end driving. Finally, the module is tested on real scenarios with real recorded traffic data, both provided by the inD dataset~\cite{inDdataset}, showing that the system is able to handle intersections unseen during the training phase and facing realistic traffic conditions.

\section{RELATED WORKS}
\label{related}
Many works handle the problem of crossing intersection using Reinforcement Learning (RL)~(\cite{rl}) algorithms. Some techniques aim to mitigate traffic congestions through reservation-based systems~(\cite{rbs}) or intelligent traffic lights~(\cite{itl1},~\cite{itl2},~\cite{itl3}). In~\cite{emergent}, authors propose a multi-agent approach in which road rules emerge as optimal solution to the traffic flow problem. Finally, in~\cite{intersection1} and~\cite{intersection2} a single agent is trained to handle unsignalized intersections using Deep Q-Networks (DQNs)~\cite{dqn}; in such works, the environment is populated by traffic cars that follow the Intelligent Driver Model (IDM)~\cite{idm} to control their speeds, while the trained vehicle chooses a discrete action in a right time step to avoid a collision with traffic cars.

Instead, our system is based on a multi-agent approach such that vehicles implicitly learn to negotiate among them as the actions of an agent affect the state of other learners and viceversa. Moreover, our agents learn to understand the priorities of all the vehicles involved in the environment, while in~\cite{intersection1} and~\cite{intersection2} it is assumed that the learner has always the lowest priority. The system is trained using a delayed version of \textit{Asynchronous Advantage Actor-Critic} (A3C,~\cite{a3c}) called Delayed-A3C (D-A3C) also used in~\cite{simtoreal} and~\cite{iri} in which has been proved that this algorithm reaches better performances than A3C for this kind of task. In particular, each agent begins the episode with a local copy of the last global network version and the system gathers the contributions of each learner during the whole episode; at a fixed time interval the agent updates its local copy of network parameters, but all the updates are sent to the global network only at the end of the episode. Instead, in A3C this exchange is performed at fixed time intervals. 

In this paper, we propose a multi-agent version of D-A3C, training all the agents in the scenario simultaneously predicting acceleration and steering angle; indeed, we do not use traffic cars regulated by rule-based methods as in~\cite{intersection1} and~\cite{intersection2} or intelligent traffic as developed in~\cite{mts}. However, we also show that such module is able to perform the maneuver in real environments populated by real traffic vehicles (Section~\ref{test_real}), proving that the proposed module does not overfit on the training scenarios both for the end-to-end driving and for the crossing maneuver tasks.

Another approach based on end-to-end driving using DRL is developed by~\cite{kendall}, where a policy for lane following was obtained training a neural network that receives a single monocular image as input; however, such model is suited to solve simple tasks and does not consider the interaction with other vehicles since it is trained in an obstacle-free environment. Other methods are based on a supervised approach as \textit{Imitation} and \textit{Conditional Imitation Learning}~(\cite{il1},~\cite{il2}); nevertheless, these techniques have the main limitation of requiring a huge amount of real data in order to avoid overfitting.

\section{NOTATION}
\subsection{Reinforcement Learning}
A typical Reinforcement Learning algorithm involves the interaction between an agent and the environment. At time $t$ the agent performs an action $a_t$ in the state $s_t$ and receives a reward signal $r_{t}$ that is typically a numerical value, and as a consequence of such action it will find itself in another state $s_{t+1}$. The RL problem can be defined as a Markov Decision Process (MDP) $M =(S, A, P, r, P_0)$, where $S$ is the set of states, $A$ the set of actions performed by the agent, $P$ the state transition probability $P(s_{t+1}|s_t, a_t)$, $r$ the reward function and $P_0$ the probability distribution of the initial state $s_0 \in S$. The goal of the RL agent is to find the best policy $\pi$ that maximizes the \textit{expected return}, defined as follows: $R_t = \sum_t^T r_t + \gamma r_{t+1} + \cdots + \gamma^{T-t}r_T$, where $T$ is the terminal state and $\gamma \in [0, 1]$ is the discount factor. 

In actor-critic methods, the model estimates two different entities: the first one is produced by the so-called \textit{Actor} and it is a probability distribution over actions ($\pi(a|s; \theta^\pi)$, where $\theta$ are the network parameters); the second one is given by the \textit{Critic} that is typically measured by the state-value function $v(s_t ; \theta^v) = \mathbb{E}(R_t|s_t)$. The updates of these functions can be defined as:
\begin{equation}
\label{theta_pi}
\theta^\pi_{t+1} = \theta^\pi_t + \alpha_\pi \cdot A_t \frac{\partial log\pi(a|s;\theta^\pi)}{\partial \theta^\pi} 
\end{equation}
\begin{equation}
\label{theta_vi}
\theta^v_{t+1} = \theta^v_t + \alpha_v \cdot A_t \frac{\partial v(s_t; \theta^v)}{\partial \theta^v}
\end{equation}
where $\alpha_\pi$ and $\alpha_v$ are the learning rates and $A_t$ is the \textit{Advantage} function defined as:
\begin{equation}
A_t = R_{t} + v(s_{t+1}; \theta^v) - v(s_t; \theta^v)
\end{equation}

In the next section we will give further details about the environments in which agents are trained.

\subsection{Environment}
\label{sec:env}

\begin{figure*}[h]
\vspace{0.3cm}
\centering
  \begin{subfigure}{.115\linewidth}
    \centering
    \includegraphics[width =\linewidth]{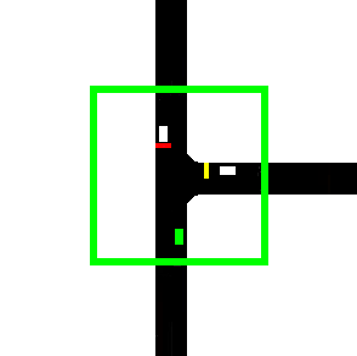}    
  \end{subfigure}
  \begin{subfigure}{.115\linewidth}
    \centering
    \includegraphics[width =\linewidth]{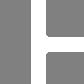}
  \end{subfigure}
  \begin{subfigure}{.115\linewidth}
    \centering
    \includegraphics[width =\linewidth]{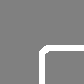}
  \end{subfigure}
  \begin{subfigure}{.115\linewidth}
    \centering
    \includegraphics[width =\linewidth]{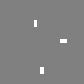}
  \end{subfigure}
  \begin{subfigure}{.115\linewidth}
    \centering
    \includegraphics[width =\linewidth]{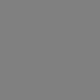}
  \end{subfigure}
  
  \vspace{0.1cm}
  
  
  \centering
  \begin{subfigure}{.115\linewidth}
    \centering
    \includegraphics[width =\linewidth]{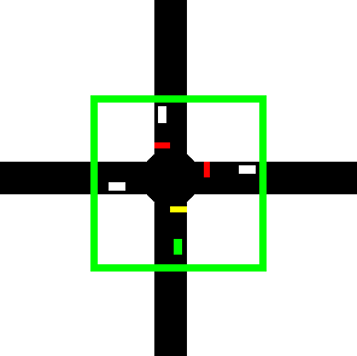}    
  \end{subfigure}
  \begin{subfigure}{.115\linewidth}
    \centering
    \includegraphics[width =\linewidth]{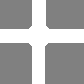}
  \end{subfigure}
  \begin{subfigure}{.115\linewidth}
    \centering
    \includegraphics[width =\linewidth]{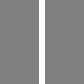}
  \end{subfigure}
  \begin{subfigure}{.115\linewidth}
    \centering
    \includegraphics[width =\linewidth]{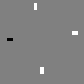}
  \end{subfigure}
  \begin{subfigure}{.115\linewidth}
    \centering
    \includegraphics[width =\linewidth]{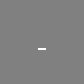}
  \end{subfigure}
  
  \vspace{0.1cm}
  
  \centering
  \begin{subfigure}{.115\linewidth}
    \centering
    \includegraphics[width =\linewidth]{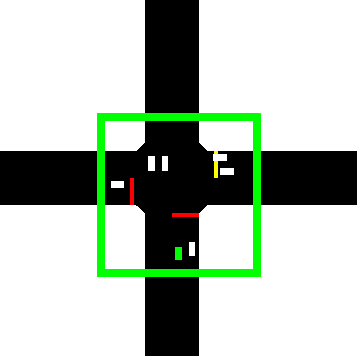}    
    \caption{Simulator}
    \label{fig:global_sim}
  \end{subfigure}
  \begin{subfigure}{.115\linewidth}
    \centering
    \includegraphics[width =\linewidth]{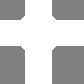}
    \caption{Map}
    \label{fig:nav_space}
  \end{subfigure}
  \begin{subfigure}{.115\linewidth}
    \centering
    \includegraphics[width =\linewidth]{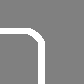}
    \caption{Path}
    \label{fig:path}
  \end{subfigure}
  \begin{subfigure}{.115\linewidth}
    \centering
    \includegraphics[width =\linewidth]{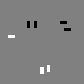}
    \caption{Obstacles}
    \label{fig:obs}
  \end{subfigure}
  \begin{subfigure}{.115\linewidth}
    \centering
    \includegraphics[width =\linewidth]{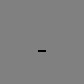}
    \caption{Traffic sign}
    \label{fig:stopline}
  \end{subfigure}
 
  \caption{Examples of views perceived by learners. The first column (Fig.~\ref{fig:global_sim}) shows the global views of the simulator, while Fig.~\ref{fig:nav_space}, Fig.~\ref{fig:path}, Fig.~\ref{fig:obs} and Fig.~\ref{fig:stopline} represent the information contained in the green squares: navigable space (map), path, obstacles including the ego-vehicle (green cars in Fig.~\ref{fig:global_sim}) and the traffic sign of the green agents respectively. The traffic sign could be black or white representing stop line and yield line respectively; if it is not present it means that no traffic sign is present, as if that lane belonged to the main road. Also the obstacles could be black, in case of those cars with higher priority than the ego-vehicle, or white representing both the ego-agent and those ones which should give the right of way to the ego-car.
  }
  \label{fig:example_views}
\end{figure*}

The scenarios used in this work are illustrated in Fig.~\ref{fig:synthetic_intersections}, corresponding to three different difficulty levels: \textit{easy} (Fig.~\ref{fig:pharmacy}), \textit{medium} (Fig.~\ref{fig:cross4_sl}) and \textit{hard} (Fig.~\ref{fig:cross4}). The first two levels are single-lane crosses, while the \textit{hard} scenario is a double-lane intersection; however, in the \textit{hard} environment we do not allow lane changes, meaning that if the agent goes out of its path we consider the episode ended as a failure. Only one agent can start from each entry lane each episode, such that the maximum number of vehicles involved simultaneously in the environment is equal to the number of entry lanes of the scenarios: 3 cars for the \textit{easy} intersection, 4 for the \textit{medium} and 8 for the \textit{hard} one. A new path and a new traffic sign are randomly assigned to the agent at the beginning of the episode, such that different state configurations are explored by the agents. However, since the \textit{hard} scenario is a double-lane intersection, we assume that both lanes of the same branch have the same traffic sign and for this reason we randomly change the traffic signs at fixed time intervals. In this way, we restart the episodes of all the learners in such scenario simultaneously, avoiding changing the traffic signs configuration during the execution of the crossing maneuver.

\begin{figure*}[h]
\vspace{0.3cm}
  \centering
  \includegraphics[width=.8\linewidth]{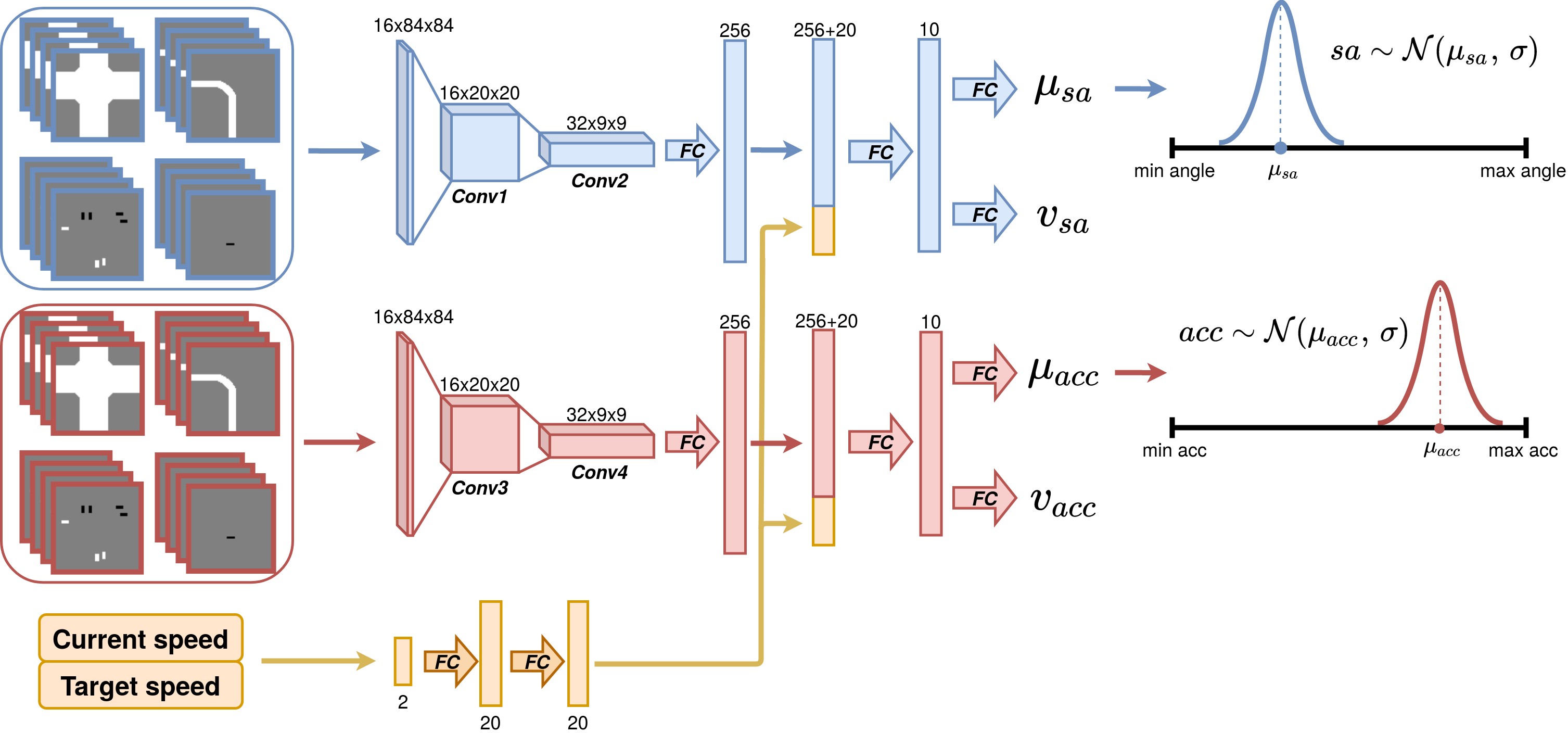}
  \caption{Neural network architecture used to train agents to navigate and cross the intersection safely. The net is composed by two different sub-modules that receive 4 visual input (navigable space, path, obstacles and traffic sign) and two scalar parameters (current speed and target speed). Each visual input is composed by 4 images 84$\times$84 pixels in order to give the agent a history of the environment to better understand the dynamic of the current state. The neural network outputs the means of two Gaussian distributions ($\mu_{acc}$ and $\mu_{sa}$), along with their state-value estimations ($v_{acc}$ and $v_{sa}$). Finally, the acceleration and steering angle are sampled using $\sigma$ that decreases linearly from 0.65 to 0.05 during the training phase.}
\label{fig:nn}
\end{figure*}

Each agent in the scenario perceives a surrounding view of 50$\times$50 meters, which is split over four different semantic images consisting on navigable space, path, obstacles and traffic sign (Fig.~\ref{fig:example_views}). These grayscale images represent one of the input of the neural network explained in Section~\ref{sec:neural_net}. Given the narrow views perceived by agents (50$\times$50 meters), we cannot handle larger intersections and for this reason future works are directed towards the implementation of a more powerful encoder (e.g. Variational Autoencoder~\cite{vae}) in order to achieve a satisfactory compression of a larger area.

Traffic signs and obstacles may assume different values; the traffic sign could represent a:
\begin{itemize}
\item Stop sign: in which case it will be drawn as a red segment in the simulator and as a black segment in the traffic sign view~(Fig.~\ref{fig:stopline}); 
\item Yield sign: it is drawn as yellow segments in the scenario and white segments in the traffic sign channel~(Fig.~\ref{fig:stopline});
\item None: no traffic sign is provided as if such lane was the main road; it means that the agent has the priority on vehicles with the stop or yield sign; however, it must always give the priority to those cars without traffic signs coming from its right (as expected by the priority to the right rule).
\end{itemize}

Obstacle views provide information about both the position of the cars inside the 50$\times$50 meters surrounding (including the ego vehicle) and the priority levels of such agents. Indeed, these obstacles can be drawn as white, as in case of the ego-car or those agents on which the ego vehicle has the right of way, or black, that are those agents with the priority over the ego-vehicle. This prior knowledge is embedded in the obstacle channel (Fig.~\ref{fig:obs}) based on both the traffic signs and the priority to the right rule (Fig.~\ref{fig:example_views}). On board of a real self-driving car, perception systems and high-definition maps could provide all the information embedded in the proposed views (Fig.~\ref{fig:example_views}), related to the road topology and vehicle priorities based on both traffic signs and the right of way rule. An example of how this kind of system can be deployed in real self-driving vehicles is shown in~\cite{simtoreal}.


Finally, we avoid cases in which all the lanes in the scenario have the same traffic sign since there could be no agent with the highest priority, but each car should give the right of way to the one on its right.

\section{THE ALGORITHM}
\subsection{Training settings}
\label{settings}
The goal of the agents is to handle the intersections safely driving along their paths through the training of a neural network that outputs both the acceleration and the steering angle every 100 milliseconds. The trained vehicles follow the kinematic bicycle model~\cite{bicycle} using values between $-0.2$ and $0.2$ for the steering angle and $-3 \frac{m}{s^2}$ and $3 \frac{m}{s^2}$ for the acceleration. Each car begins the episode with different speeds randomly picked inside the interval $[3.0, 6.0]\frac{m}{s}$ and different target speed ($[7.0, 10.0]\frac{m}{s}$) that is the maximum velocity the vehicle should not exceed. Since the traffic is represented by the learners, each agent waits a random delay $[delay_{min}, delay_{max}]$ before starting a new episode in order to ensure variability in the traffic density. This delay range depends on the scenario and could assume different values: $[0, 30]$, $[0, 50]$ and $[0, 100]$ seconds for the \textit{easy}~(Fig.~\ref{fig:pharmacy}), \textit{medium}~(Fig.~\ref{fig:cross4_sl}) and \textit{hard}~(Fig.~\ref{fig:cross4}) intersection respectively. 

Finally, the episode could end due to four different events: 
\begin{itemize}
\item the agent reaches the goal position crossing the intersection safely and driving without going out of its path;
\item the vehicle crashes with another vehicle;
\item the agent goes out of its path;
\item the time available to finish the episode expires.
\end{itemize}

\subsection{Neural Network Architecture}
\label{sec:neural_net}

Agents are trained using a neural network composed by two sub-modules: one to handle the acceleration (\textit{acc}) and the other one to manage the steering angle (\textit{sa}). As shown in Fig.~\ref{fig:nn}, the sub-modules receive a sequence of four visual input (Fig.~\ref{fig:nav_space},~\ref{fig:path},~\ref{fig:obs}, and~\ref{fig:stopline}), each one composed by four images 84$\times$84 pixels. Along with this visual input, the neural network receives two scalar parameters representing the agent speed and the target speed of the agent. In order to ensure exploration, the actions performed by the agent are sampled by two Gaussian distribution centered on the output of the two sub-modules ($\mu_{acc}$,~$\mu_{sa}$). In this way the acceleration and the steering angle can be defined as: ${acc\sim\mathcal{N}(\mu_{acc},\,\sigma)}$ and $sa \sim \mathcal{N}(\mu_{sa},\,\sigma)$, where $\sigma$ is a tunable parameter and it decreases linearly from 0.65 to 0.05 during the training phase. Together with $\mu_{acc}$ and $\mu_{sa}$, the neural network produces the corresponding state-value estimations ($v_{acc}$ and $v_{sa}$) using two different reward functions: $R_{acc, t}$ and $R_{sa, t}$. These signals are related to the acceleration and steering angle output respectively, such that the state-value functions can be written as: ${v_{acc}(s_t;\theta^{v_{acc}}) = \mathbb{E}(R_{acc, t}|s_t)}$ and $v_{sa}(s_t;\theta^{v_{sa}}) = \mathbb{E}(R_{sa, t}|s_t)$.

In this case, the policy update described in Equation~(\ref{theta_pi}) can be defined as follows:
\begin{align}
\theta_{t+1}^{\mu_{acc}} = \theta_{t}^{\mu_{acc}} + \alpha \cdot A_{acc, t} \frac{\nabla \pi(a_t|s_t; \theta_{t}^{\mu_{acc}})}{\pi(a_t|s_t; \theta_{t}^{\mu_{acc}})} = \nonumber\\ 
\theta_{t}^{\mu_{acc}} + \alpha \cdot A_{acc, t} \frac{\nabla \mathcal{N}(acc|\mu_{acc}(\theta_t^{\mu_{acc}}))}{\mathcal{N}(acc|\mu_{acc}(\theta_t^{\mu_{acc}}))} = \nonumber\\
\theta_{t}^{\mu_{acc}} + \alpha \cdot A_{acc, t} \frac{acc - \mu_{acc}}{\sigma^2} \nabla \mu_{acc} (\theta_t^{\mu_{acc}})
\end{align}
and the \textit{Advantage} ($A_{acc, t}$) as:
\begin{equation}
A_{acc, t} = R_{acc, t} + v_{acc}(s_{t+1}; \theta^{v_{acc}}) - v_{acc}(s_t; \theta^{v_{acc}})
\end{equation}
The same equations can be written for the steering angle output, replacing $\mu_{acc}$ with $\mu_{sa}$ and $acc$ with $sa$.


\subsection{Reward Shaping}
We defined two different reward functions $R_{acc, t}$ and $R_{sa, t}$ in order to evaluate the acceleration and the steering angle output separately. As explained in Section~\ref{sec:env}, we do not allow lane changes and for this reason we assume that a crash between two vehicles and the off-road case just depends on the acceleration and steering angle output respectively. $R_{acc, t}$ and $R_{sa, t}$ could be defined as follows:
\begin{equation}
R_{acc, t} = r_{speed} + r_{terminal}
\end{equation}
\begin{equation}
R_{sa, t} =  r_{localization} + r_{terminal}
\end{equation}

$r_{speed}$ is a positive reward related to $R_{acc, t}$ in order to encourage the agent to reach the \textit{target speed} and it is defined as:
\begin{equation}
r_{speed} = \xi \cdot \frac{current\ speed}{target\ speed}
\end{equation}
where $\xi$ is a constant set to 0.005.

$r_{terminal}$ is present both in $R_{acc, t}$ and $R_{sa, t}$ and it could assume different values depending on the terminal state of the episode, which can be one of the following:
\begin{itemize}
\item Goal reached: the agent crosses the intersection safely, without going outside its path and without crashing with another vehicle. The value used for $r_{terminal}$ is $+1$ both for $R_{acc, t}$ and $R_{sa, t}$.
\item Off-road: the episode ends when the agent goes off its path and we assume that it is due only to an inaccurate estimation of the steering angle output. For this reason $r_{terminal}$ will be $0$ for $R_{acc, t}$ and $-1$ for $R_{sa, t}$.
\item Crash: the episode ends with a crash with another vehicle and we assume that this is due only to an inaccurate estimation of the acceleration output since the lane change is not allowed and it would generate the \textit{Off-road} terminal state. For this reason $r_{terminal}$ will be $0$ for $R_{sa, y}$. In case of $R_{acc, t}$, we modulate its value based on the coding explained in~Section~\ref{sec:env}, by which the ego car should give the right of way to black agents, but it has the priority on white vehicles~(Fig.~\ref{fig:obs}). By encoding such information inside the obstacle channel, we penalize the cars involved in the accident setting the value of $r_{terminal}$ to $-1$ in the case the ego-agent commits an accident with a black car, otherwise it will be $-0.5$.
\item Time over: the time available to end the episode expires and this is closely related to a conservative acceleration profile; for this reason, $r_{terminal}$ will assume the value of $0$ for $R_{sa, t}$ and $-1$ for $R_{acc, t}$.
\end{itemize}

Finally, $r_{localization}$ is a penalization related to $R_{sa, t}$ given to the agent when its position ($x, y$) and its heading ($h_a$) differs from the center lane and the heading ($h_p$) of the path respectively. This factor can be defined as:
\begin{equation}
r_{localization} = \phi \cdot \cos (h_a - h_p) + \psi \cdot d
\end{equation}
where $\phi$ and $\psi$ are constants set to 0.05 and $d$ is the distance between the position of the agent and the center of its path.

The reward shaping illustrated in this section is essential to obtain a module able to navigate the vehicle safely, learning the right of way rule in order to handle intersection scenarios.

\begin{table*}[h]
\vspace{0.3cm}
\centering
  \caption{Results obtained in the training scenarios (Fig.~\ref{fig:synthetic_intersections}) with the three different traffic signs.}
  \label{tab:baseline}
  \begin{tabular}{|r|c|c|c||c|c|c||c|c|c|}
  	\hline
  		 & \multicolumn{3}{c||}{Easy Intersection} & 
   		   \multicolumn{3}{c||}{Medium Intersection} & 
   		   \multicolumn{3}{c|}{Hard Intersection} \\ 
    \hline
    	& No Sign & Yield Sign & Stop Sign & No Sign & Yield Sign &	Stop Sign & No Sign & Yield Sign & Stop Sign \\  
    \hline
    	Reaches \% & 0.998 & 0.995 & 0.992 & 0.995 & 0.991 & 0.992 & 0.997 & 0.991 & 0.972 \\
    	Crashes \% & 0.002 & 0.005 & 0.008 & 0.005 & 0.009 & 0.008 & 0.003 & 0.009 & 0.012 \\
    	Off-roads \% & 0.0 & 0.0 & 0.0 &  0.0 & 0.0 & 0.0  & 0.0 & 0.0 & 0.0 \\
    	Time-overs \% & 0.0 & 0.0 & 0.0 &  0.0 & 0.0 & 0.0  & 0.0 & 0.0 & 0.016 \\
    	Average Speed $[\frac{m}{s}]$ & 8.533 & 8.280 & 8.105 & 8.394 & 7.939 & 7.446  & 8.244 & 7.365 & 5.855 \\ 
    \hline
  \end{tabular}
\end{table*}

\begin{figure*}[b]
  \centering
  \begin{subfigure}{.239\linewidth}
    \centering
    \includegraphics[width =\linewidth]{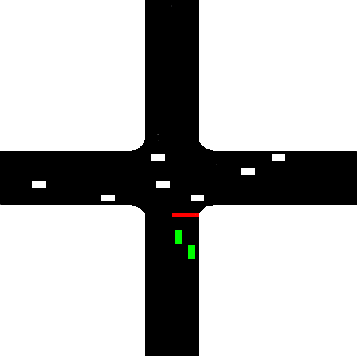}
    \caption{Test scenario}
    \label{fig:ttc}
  \end{subfigure}
  \begin{subfigure}{0.28\linewidth}
    \centering
    \includegraphics[width =\linewidth]{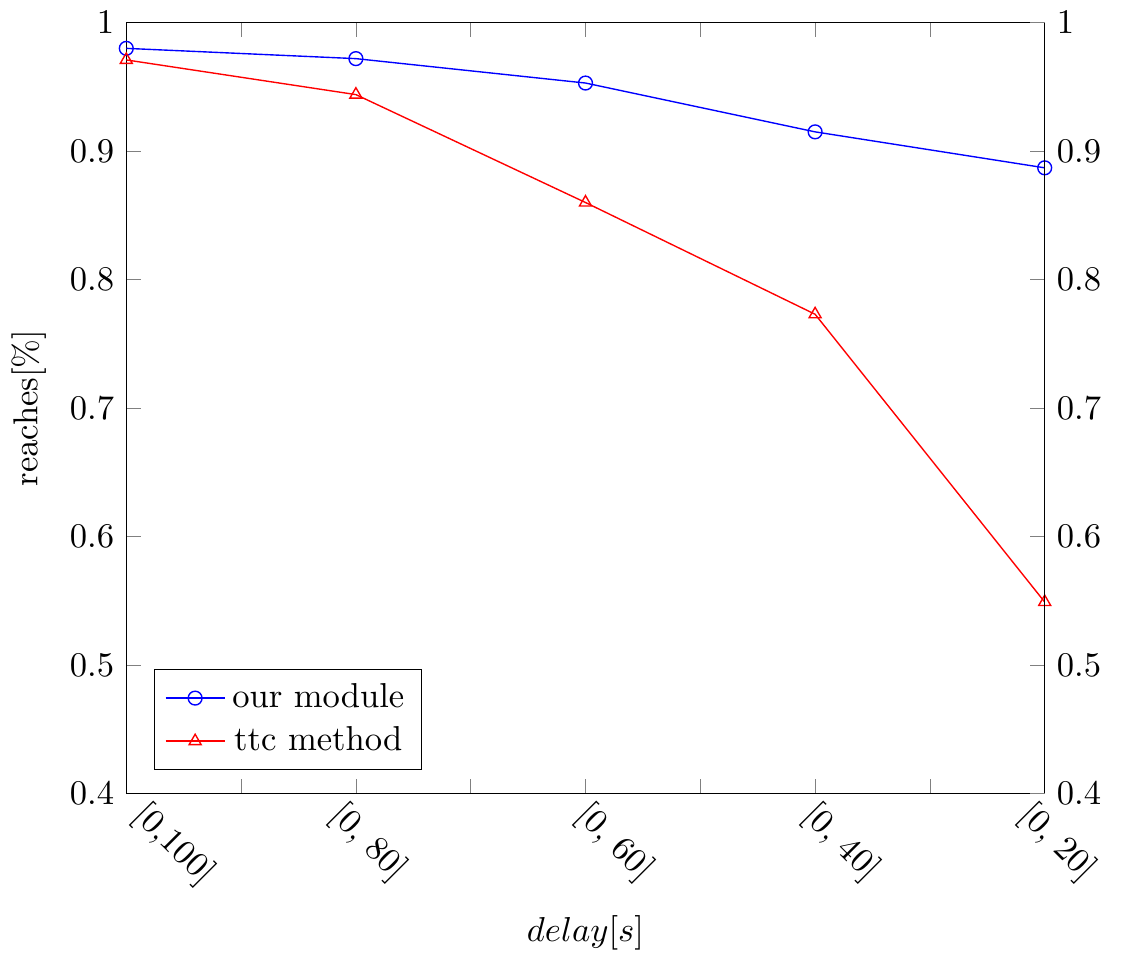}    
    \caption{Reaches}
    \label{fig:ttc_reaches}
  \end{subfigure}
  \begin{subfigure}{0.28\linewidth}
    \centering
    \includegraphics[width =\linewidth]{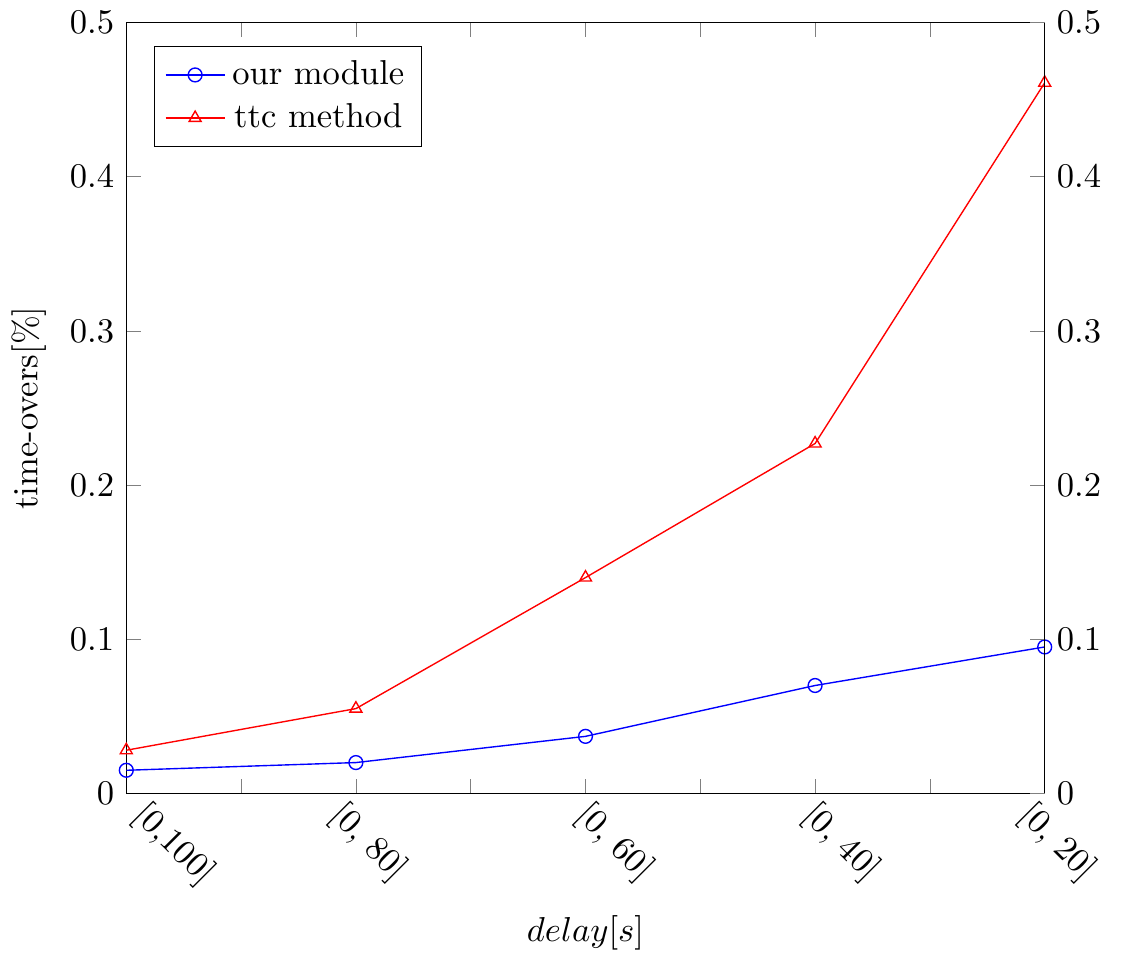}
    \caption{Time-overs}
    \label{fig:ttc_time}
  \end{subfigure}
  \caption{Fig.~\ref{fig:ttc} shows an example of the environment used to compare the performance of our module with the time-to-collision (TTC) algorithm. Green agents are those vehicles that perform the crossing maneuver, while the white ones represent traffic cars that follow the path using the Intelligent Driver Model (IDM) to control their speeds. Fig.~\ref{fig:ttc_reaches} and Fig.~\ref{fig:ttc_time} show the comparison between the percentages of episodes ended successfully and the percentage of time-overs obtained by our module (blue) and the time-to-collision (TTC) algorithm (red) respectively.}
  \label{fig:ttc_test}
\end{figure*}

\section{EXPERIMENTS}
\label{experiments}
In this section we will show how agents learn successfully both to navigate along their paths and to cross intersections safely. We test the system on the training scenarios with different traffic conditions and traffic signs configurations; as explained in Section~\ref{related}, it is difficult to compare our results with those obtained in similar previous works as~\cite{intersection1} and~\cite{intersection2}, since they focus on training a single agent that does not really learn the right of way rule, but it only chooses the right time to perform discrete actions avoiding collisions with traffic vehicles. Moreover, these works do not manage different kind of intersections with different traffic signs assuming that the learner has always the lowest priority in the intersection. Instead, in our multi-agent approach agents handle intersection scenarios learning the right of way rule based on the traffic sign and on the priority to the right rule; this is essential to achieve a human-like behavior and to avoid unnecessary delays. Indeed, we also compare our module with the time-to-collision algorithm (TTC) showing that our system reaches better results than this rule-based method. Finally, we show that the proposed system is able to drive in environments unseen during the training phase, performing the crossing maneuver facing real traffic vehicles.

\subsection{Results}
\label{sec:baseline}
We analyze the results achieved on each intersection (Fig.~\ref{fig:synthetic_intersections}) based on traffic signs and the metrics used for this test are: \textit{Reaches}, \textit{Crashes}, \textit{Off-roads} and \textit{Time-overs} corresponding to the percentage of episodes ended successfully, with a crash between two vehicles, with the agent off its path and due to the depletion of available time respectively. Moreovoer, we also analyze a further metric, that is the average speed of the agents during the episodes.

For each scenario we performed several experiments analyzing the agent behavior facing the three different traffic signs. Each test is composed by 3000 episodes using different traffic conditions. Indeed, the traffic configuration changes based on a random delay $[delay_{min}, delay_{max}]$ that vehicles have to wait before starting a new episode: the lower the delay range, the heavier the traffic condition. The values of $delay_{min}$ and $delay_{max}$ are set to 0 and 10 seconds respectively and we performed as many tests as $delay_{max}$ reaches the maximum delay used during the training phase increasing its value by 10 seconds for each test. Recalling that the values of $delay_{max}$ used during the training phase are 30, 50 and 100 seconds for the three scenarios, we performed three experiments for the \textit{easy intersection} (Fig.~\ref{fig:pharmacy}), five tests for the \textit{medium} (Fig.~\ref{fig:cross4_sl}) and ten for the \textit{hard} one (Fig.~\ref{fig:cross4}).

The results illustrated in Table~\ref{tab:baseline} represent the average percentages of these experiments. As we can notice, crossing the intersection with the stop sign is the hardest task since the agent has often the lowest priority. Moreover, observing the \textit{Average Speed} values, we can notice how agents modulate their behaviors based on the traffic sign in order to achieve the goal safely. Finally, we can notice that the off-road case never happens in all the experiments, proving that agents learn to drive safely along their paths.

The following video\footnote{\href{https://youtu.be/x28qRJXiQfo}{https://youtu.be/x28qRJXiQfo}} shows the behavior of the agents in the three training scenarios with different traffic signs.


\begin{figure*}[h]
  \centering
  \begin{subfigure}{.204\linewidth}
    \centering
    \includegraphics[width =\linewidth]{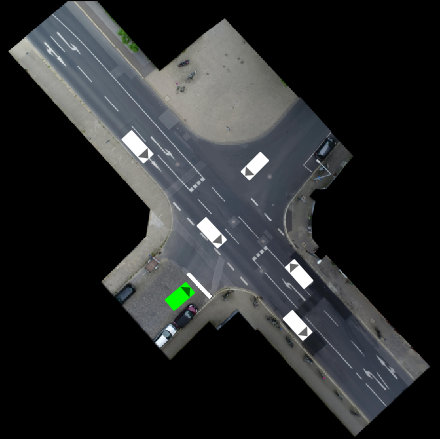}    
    \label{fig:real_1}
  \end{subfigure}
  \begin{subfigure}{.204\linewidth}
    \centering
    \includegraphics[width =\linewidth]{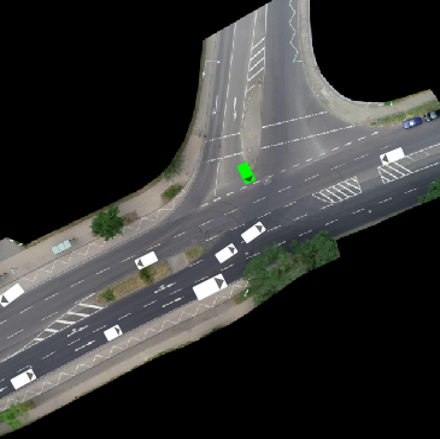}
    \label{fig:real_2}
  \end{subfigure}
  \begin{subfigure}{.204\linewidth}
    \centering
    \includegraphics[width =\linewidth]{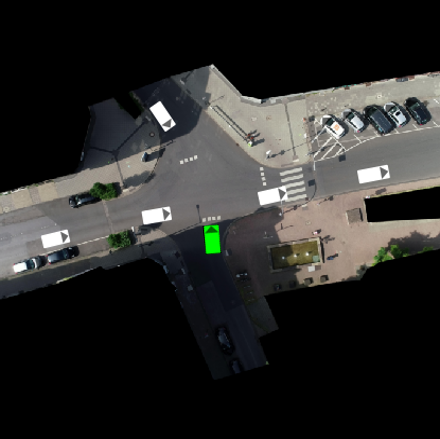}
    \label{fig:real_3}
  \end{subfigure}
  \begin{subfigure}{.204\linewidth}
    \centering
    \includegraphics[width =\linewidth]{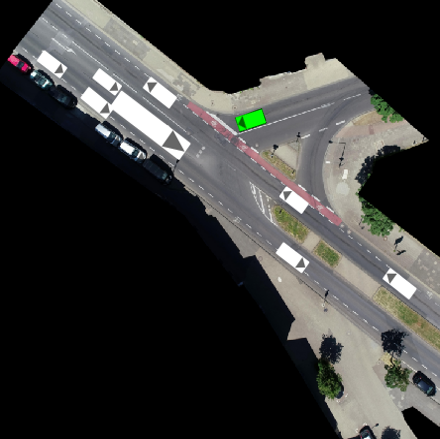}
    \label{fig:real_4}
  \end{subfigure}
  \caption{Real scenarios of the inD dataset~\cite{inDdataset} used for testing our module. Green cars represent our RL agents, while recorded road users (cars, trucks and busses) are illustrated as white vehicles. }
  \label{fig:real_envs}
\end{figure*}

\subsection{Comparison with TTC method}
In this test we compare our module with the rule-based method of time-to-collision (TTC). We use the \textit{hard intersection} and we let the agent performing the crossing maneuver from the branch with the stop sign as illustrated in Fig.~\ref{fig:ttc}. Traffic cars start from the two adjacent branches of the ego-vehicle and they follow the center of the path using the Intelligent Driver Model (IDM) to control their speeds.

In the TTC test, agents start the episode decelerating in order to reach the stop line with zero speed and following the center of the path without predicting the steering angle. Then, a single threshold is used to estimate when accelerate to cross the intersection: considering an imaginary line aligned with the longitudinal axis of the ego-agent (green vehicles in Fig.~\ref{fig:ttc}), the TTC is computed as the time a traffic car reaches this line, assuming that it drives with constant speed. Each time step we calculate the TTC of all the traffic cars and if the lowest value exceeds a specific threshold, the ego-agent accelerates ($3\frac{m}{s^2}$), otherwise it continues waiting at the stop line. This threshold is chosen in order to obtain the best results with the TTC method and zero percent of \textit{Crashes}. 

The test is performed using different traffic condition levels in which both our module and TTC method have the same available time to finish the episode. The traffic levels are: \textit{low}, \textit{medium} and \textit{high}, corresponding to a maximum number of traffic cars involved in the environment to 4, 8 and 12 respectively. For each configuration, traffic agents wait a random delay before starting a new episode as explained in Section~\ref{sec:baseline} in order to gradually increase the difficulty of the crossing maneuver; for these experiments, $delay_{max}$ is increased by 20 seconds for each test. We noticed that for \textit{low} and \textit{medium} levels of traffic conditions both the TTC method and our module achieve good results (more than $99\%$ of episodes ended successfully); however, the TTC method drops in the case of \textit{high} traffic condition, leading the system to undefined waits. The comparison between our model and TTC method is illustrated in Fig.~\ref{fig:ttc_reaches} and Fig.~\ref{fig:ttc_time}, showing that the rule-based method leads the system to unnecessary delays. The difference between TTC and our system becomes more and more evident increasing the traffic level of the environment, namely reducing the delay the traffic cars have to wait before starting a new episode: the smaller the delay, the greater the difficulty to cross the intersection since more traffic agents will be involved in the scenario simultaneously.

\begin{table}[b]
  \caption{Percentages of episodes ended following the right of way rule by all the agents involved in the episode (\textit{No infraction}) and with at least an infraction (\textit{Infraction}).}
  \label{tab:synchro}
  \begin{center}
  \begin{tabular}{|r|c|c|c|}  
  	\hline
    	& Easy & Medium & Hard \\ 
    	& Intersection & Intersection & Intersection \\   
	\hline    
    	No infraction \% & 0.985 & 0.990 & 0.863 \\ \hline
    	Infraction \% & 0.015 & 0.010 & 0.137 \\ 
    \hline
  \end{tabular}
  \end{center}  
\end{table}

\subsection{Testing the Right of Way Rule}
As explained in Section~\ref{sec:env}, agents handle the intersection learning the priority to the right rule and we proved that they learn to drive and handle the intersection scenarios safely (Section~\ref{sec:baseline}). However, the results obtained in Table~\ref{tab:baseline} do not prove that agents respect the right of way rule. 

At this purpose, we propose a test on the training scenarios (Fig.~\ref{fig:synthetic_intersections}) in which agents start the episode synchronously with a random path and traffic sign, setting the same current and target speed for all cars, such that vehicles approach the intersection almost simultaneously. As for the training phase, only one agent can start from each lane such that the maximum number of vehicles involved in each episode is 3, 4 and 8 for the \textit{easy}, \textit{medium} and \textit{hard} intersection respectively. Only in this way, we can understand if an agent crosses the intersection before a vehicle with higher priority, thus committing an infraction. For this test we consider the episode ended when all the agents involved in the scenario reach a terminal state.

Each test on the three intersections (Fig.~\ref{fig:synthetic_intersections}) is composed by 9000 episodes and Table~\ref{tab:synchro} shows the percentages of episodes ended successfully with and without infractions. We consider the episode ended without infractions (\textit{No Infraction}) if all vehicles involved in the scenario respect the right of way rule, otherwise the episode will be considered as a failure (\textit{Infraction}). We can notice that agents respect the rule in most of the cases and only in the \textit{hard intersection} there is a slight decrease in the performances since the larger spaces of such environment allow vehicles to hazard the maneuver more frequently.

\subsection{Test on Real Data}
\label{test_real}

We test our module using the inD dataset~(\cite{inDdataset}), containing 33 real traffic sequences in four different intersections (Fig.~\ref{fig:real_envs}) for a total of 10 hours of recorded data. The dataset is collected using a camera-equipped drone and it contains the tracking and classification of more than 11500 static and dynamic road users including cars, trucks, busses, bicyclists and pedestrians; however, we consider only dynamic vehicles data of cars, trucks and busses, since pedestrian and bicyclist obstacles are not included in our simulator in the training phase. During these tests, we assume that our RL agents (green cars in Fig.~\ref{fig:real_envs}) have always the lowest priority since they are not sensed by traffic vehicles. The RL agent always starts from the same lane (the least busy), choosing a random exit each episode; for this reason, we do not include those traffic vehicles starting from the same lane of the RL agent. In this way, the total number of traffic vehicles populating the four real scenarios is 7386, while the episodes performed by our RL agents are 2702.

Using the recorded trajectories of the dynamic traffic vehicles, we built the four environments (Fig.~\ref{fig:real_envs}) in our simulator with CAIRO graphic library~\cite{cairo}. For each episode, we saved data related to the RL agent (position, speed and heading) in order to project them on the real scenarios (Fig.~\ref{fig:real_envs}) using the code provided by~\cite{inDdataset} on their github page\footnote{\href{https://github.com/ika-rwth-aachen/drone-dataset-tools}{https://github.com/ika-rwth-aachen/drone-dataset-tools}}. 

The following video\footnote{\href{https://youtu.be/SnKUk2k9YCg}{https://youtu.be/SnKUk2k9YCg}} shows how the RL agents drive and perform the crossing maneuver in some recorded sequences of such real scenarios. The percentage of episodes ended successfully is greater than $99\%$, with $0\%$ time-over and $0\%$ off-road cases. Moreover, some inaccurate evaluation in the maneuver execution are emphasized as the RL agent is not sensed by traffic vehicles. However, considering the diversity of scenarios and traffic behavior between training and test phase, we can state that this test represents a promising result for future tests in real-world environments, in which such module should always be flanked with safety systems in order to avoid collisions. In addition, a smoother driving style should be achieved in future works in order to test such system on board of a real self-driving vehicle.

Finally, the results obtained in this test together with the behavior of the RL agent illustrated in the video, prove that our module is able to generalize both to unseen scenarios and to different real traffic conditions.

\section{CONCLUSION}
In this work we presented a system able to handle intersection scenarios regulated by traffic signs in which vehicles have to understand the priorities of other cars involved in the environment following the priority to the right rule. Agents are trained using a neural network that outputs both acceleration and steering angle at each time step. We proved that vehicles learn the right of way rule encoding the only information on agents priorities in the obstacle channel (Fig.~\ref{fig:obs}); in this way, each agent observes two types of vehicles, learning which cars have higher priorities and which ones should let it pass. We reached good performances on three different intersection environments (Fig.~\ref{fig:synthetic_intersections}), comparing our system with the time-to-collision (TTC) algorithm showing that increasing the traffic level, our module reached better results than the rule-based method that leads the system to unnecessary delays. Finally, we also proved that the proposed system is able to generalize both to unseen scenarios and to real traffic conditions. The results show that our module is able to perform the maneuver in presence of different human driving behaviors even if the system was fully trained in a multi-agent fashion without real traffic data.

\bibliography{root} 
\bibliographystyle{ieeetr}

\end{document}